# Influence of Control Parameters and the Size of Biomedical Image Datasets on the Success of Adversarial Attacks

Vassili Kovalev[1] and Dmitry Voynov[2]

[1] United Institute of Informatics Problems, Surganova St, 7, 220012 Minsk, Belarus
[2] Belarus State University, Nezavisimosti Av, 4, 220030 Minsk, Belarus
{vassili.kovalev, voynovdd}@gmail.com

**Abstract.** In this paper, we study dependence of the success rate of adversarial attacks to the Deep Neural Networks on the biomedical image type, control parameters, and image dataset size. With this work we are going to contribute towards accumulation of experimental results on adversarial attacks for the community dealing with biomedical images. The white-box Projected Gradient Descent attacks were examined based on 8 classification tasks and 13 image datasets containing a total of 605,080 chest X-ray and 317,000 histology images of malignant tumors. We concluded that: (1) An increase of the amplitude of perturbation in generating malicious adversarial images leads to a growth of the fraction of successful attacks for the majority of image types examined in this study. (2) Histology images tend to be less sensitive to the growth of amplitude of adversarial perturbations. (3) Percentage of successful attacks is growing with an increase of the number of iterations of the algorithm of generating adversarial perturbations with an asymptotic stabilization. (4) It was found that the success of attacks dropping dramatically when the original confidence of predicting image class exceeds 0.95. (5) The expected dependence of percentage of successful attacks on the size of image training set was not confirmed.

## 1 Introduction

### 1.1 The Problem of Security of Computerized Diagnosis

It is well recognized that the security issues of computerized disease diagnosis are of paramount importance. Recently, the Deep Learning technologies gave the community well-grounded promises to become an effective tool in biomedical image analysis and computerized diagnosis [1, 2]. Currently, a large fraction of studies focused on the development of increasingly more accurate models while less attention has been given to the security and robustness of these models. Unfortunately, it was found that along with the high success, the Deep Learning brought some new security problems. This time the security worries arose from the vulnerability of methods capitalizing on Deep Neural Networks (DNN) to so-called adversarial attacks. The vulnerability to adversarial image examples was first noticed in [3] and more systematically discussed by researchers from Google in [4]. A bit later, in 2015-2016 a group of researchers has provided several examples of the vulnerability of DNNs to adversarial attacks [5]. It was experimentally proven that adversarial attacks may lead to failures in making correct classification decision. In the context of Computer Assisted Diagnosis (CAD) systems this corresponds to possible failures in correct disease diagnosis. Contrary to the "traditional" computer viruses, these attacks are performed with the help of so-called adversarial images which are nothing but specially crafted images that push DNNs to wrong classification decisions. More recently, there have been several works published on the problem of adversarial attacks, their types, and possible ways of defense (see, for example, surveys [6, 7] and paper 8).

### 1.2 Basic Properties of Adversarial Attacks and the Purpose of This Study

In general, the problem of adversarial attack is not studied well as yet. Moreover, several key points that have been discovered recently are counter-intuitive. The basic properties of adversarial attacks were intensively studied in [3, 4, 8] and illustrated on large image datasets coming from computer vision applications. In our view, the most interesting facts and statements from papers [3, 4, 8] which are relevant to the present study are as follows:

- The state-of-the-art DNNs demonstrate good generalization performance on different image classification tasks with high variability of image objects and backgrounds. Thus, we can expect such networks should be robust to small perturbations of input images. However, it is shown that very tiny and even imperceptible non-random (i.e., intentional, malicious, adversarial) image modifications could totally change the network's prediction result.
- The adversarial images which were specially crafted to attack one specific DNN are statistically hard to correctly classify for other DNNs too. This holds true even under condition the other DNNs are of different architecture, trained with different parameters and even on different image subsets.
- A popular approach in computer vision is to use convolutional network features as space where Euclidean distance approximates perceptual distance. However, this resemblance is clearly flawed if images that have an immeasurably small perceptual distance correspond to completely different classes in the network's representation.

It should be noted, though, that the majority of activities on studying adversarial attacks and possible methods of defense are carried out in the field of computer vision but not in the biomedical imaging domain.

Thus, the purpose of this paper is to study dependence of the success rate of adversarial attacks on the image type, image dataset size, and control parameters. In a more general context, with this work we are going to contribute towards accumulation of experimental materials on adversarial attacks for the biomedical imaging community.

## 2 Original Image Data

For studying adversarial attacks, we have selected two different kinds of biomedical images. The first image type is represented by chest X-ray images which often used in a screening for detecting lung diseases, diseases of cardiovascular system, and skeletal abnormalities as well as for monitoring various treatment processes. The second image type was represented by color histological images which are continuously playing the role of a gold standard in cancer diagnosis. These two opposite image types were chosen for experimentation because chest X-rays holding certain anatomical shape with the relatively high role of spatial, "geometrical" structure whereas the histological images can be viewed as a "shape-free" color texture.

### 2.1 Chest X-ray Images

**Norm of Chest X-Ray (X-Norm-All).** All X-ray image data used with this study were the natively-digital X-ray scans which were extracted from a PACS system. The PACS records contain information on the results of a periodic chest screening of the population of a two-million city conducted during the years 2001–2014. Thus, the reduced version of the database we used here contains a total of 1,908,926 records. Each record corresponds to a single digital chest X-ray image. All the records also included data on patients' age, gender, and textual radiological reports. The reports are written in a free-form native, non-English language. Each report was made by an experienced radiologist. In complicated cases these reports were resulted from a joint image reading by a small board of radiologists. All the radiologists were employed in the framework of a large-scale telemedicine screening system on a permanent basis.

The radiological reports present information on possible lung diseases, diseases of cardiovascular system (heart and blood vessels), and information on possible skeletal abnormalities such as scoliosis, deformation of ribs, etc. The list of lung abnormalities included pneumosclerosis, emphysema, fibrosis, pneumonia, focal shadows, bronchitis, and lung tuberculosis. Technically, all the X-ray scans were represented by 1-channel 16-bit non-compressed images which were originally stored in DICOM format. The image resolution varied from rarely occurred small size of 520×576 pixels to relatively large fraction which was sized to 2800×2531 pixels. Since all the images were natively digital, there were no film scanning artifacts presented in the image data.

An image was categorized into the class of Norm if there were no visible signs of any type of abnormalities in mediastinum, skeleton and the lungs themselves. Normal cases were selected from the database by way of parsing of radiological reports with the help of a small set of a limited number of key phrases the radiologists typically use to label images with the absence of any visible abnormalities. Because the well sufficient amount of images of the Norm was available, the only strict matches of the report texts with one of the key phrases have been considered. A small fraction of technically unusable images was discarded by the same reason. Following this procedure, a total of 1,215,648 cases were stored to the basic image dataset referred to as X-Norm-All which was used as a repository of the chest X-Ray of Norm in this study.

**Pathology of Chest X-Ray (X-Path-All).** Similar to the Norm, labeling of abnormal lung cases was done using a keyword match in the textual descriptions. As a result, 22,355 cases were labeled as belonging to the class of pneumosclerosis, 9,285 items were recognized as emphysema, 19,844 images were categorized to the fibrosis, 5,718 images got labels of "pneumonia", 2,897 lungs were recognized as the ones with focal shadows, 821 subjects were classified to the class of having bronchitis, and 793 cases have received the label of "tuberculosis". It should be appropriately stressed that one single case may have several labels of abnormality simultaneously. Also, it should be remembered that here we are dealing with the screening data. This particularly means that the above lung disease labels are preliminary and are not confirmed clinically. In other words, we should admit the presence of certain bias which is caused by over-estimation of the probability of presence of pathological changes what is natural for the screening stage. Finally, as a result of the above selection procedure, we end up with a total of 46,882 X-ray scans presenting various lung abnormalities.

## 2.2 Histology Images

**Ovary and Thyroid Cancer (H-OV-TH).** The test histology image dataset was acquired from 46 patients with confirmed diagnosis of Ovary cancer (23 patients) and Thyroid cancer (23 other patients). All the data were sub-sampled from a private database resulted from previous project on studying angiogenesis of malignant tumors. Tissue probes represented tumor regions and surrounding non-tumor tissue which is conditionally termed here as Norm. After biopsy the issue samples were immunohistochemically processed using D2–40 mesothelial marker for highlighting tumor angiogenesis. An intermediate image dataset consisting of 4000 images and representing 4 classes including Ovary Tumor (H-OV-T, 1000 images), Ovary Norm (H-OV-N, 1000 images), Thyroid Tumor (H-TH-T, 1000 images), and Thyroid Norm (H-TH-N) was acquired with the help of fully-digital Leica DMD108 optical microscope. Finally, to be used for benchmarking the adversarial attacks, the intermediate color images of 2048×1536 pixels in size were partitioned into 256×256 sections without overlap. This procedure resulted in the histology image dataset H-OV-TH consisting of a total of 192,000 images, 48,000 images in each of four of above abbreviated classes.

**Breast Cancer (H-BR).** Brest cancer histology images employed with this study were sampled from a dataset of 500 whole slide Hematoxylin-Eosin stained images used for benchmarking of methods of prediction of breast cancer proliferation score [9]. A total of 125,000 color images of 256×256 pixels in size representing malignant tumors (H-BR-N, 62,500 images) and areas not affected by tumors (H-BR-N, 62,500 images, conditionally referred here to as Norm) were arbitrary sampled from 500 whole slide images on the highest resolution level.

## 2.3 Study Groups

Adversarial attacks were examined with the help of dedicated study groups presented in Table 1 together with their key characteristics.

**Table 1.** Study groups and their characteristics

| Image type | Acronym | Classification task | N img total | N img by classes | CLS score |
|---|---|---|---|---|---|
| X-ray, Norm | X-NR2 | 2 age groups: G1: 20-35 years G2 50-70 years | 200,000 | G1: 100,000 G2: 100,000 | 0.98 |
| X-ray, Norm | X-NR3 | 2 age groups: G1: 17-24 years G2: 25-41 years G3: 42-80 years | 550,080 | G1: 183,360 G2: 183,360 G3: 183,360 | 0.83 |
| X-ray, region of Aorta | X-AO | 2 classes: C1: Aorta with anterior rotation C2: Norm | 27,000 | C1: 10,980 C2: 16,020 | 0.78 |
| X-ray, Tuberculosis | X-TB | 2 classes: C1: Tuberculosis C2: Norm | 28,000 | C1: 14,000 (1,369) C2: 14,000 | 0.82 |
| Histology, Ovary cancer | H-OV | 2 classes: C1: Tumor C2: Norm | 96,000 | C1: 48,000 C2: 48,000 | 0.92 |
| Histology, cancer of Thyroid Gland | H-TH | 2 classes: C1: Tumor C2: Norm | 96,000 | C1: 48,000 C2: 48,000 | 0.94 |
| Histology, Ovary cancer and cancer of Thyroid Gland | H-OV-TH | 4 classes: C1: Ovary Tumor C2: Ovary Norm C3: Thyroid Tumor C2: Thyroid Norm | 192,000 | C1: 48,000 C2: 48,000 C3: 48,000 C4: 48,000 | 0.91 |
| Histology, Breast cancer | H-BR | 2 classes: C1: Tumor C2: Norm | 125,000 | C1: 62,500 C2: 62,500 | 0.97 |

For brevity purposes the necessary explanations on creation of study groups provided below in an itemized style.

1. Groups X-NR2 and X-NR3 of chest images of Norm of subjects of different ages constitute good test bed for classification tasks due to known age-related changes. In both cases the main principles for selecting images from original dataset X-Norm-All containing 1,215,648 items was to preserve gender balance as well as equivalent population of classification sub-groups.
2. In the aorta classification tasks X-AO the areas containing anterior regions of aorta were extracted automatically from original X-ray images and final results were examined visually. Since original images were of different size, these regions re-scaled to 256×256 based on individual resolution.

3. Due to the lack of lung tuberculosis images, the available dataset of 1,369 images was inputted to a dedicated augmentation procedure to finally get 14,000 items. Note that this is *only the case* in this study where the augmentation procedure was used.
4. The last column of Table 1 presents the original classification accuracy.

In all the occasions except for classification tasks X-NR3 the images used for experimentation on adversarial attacks were 256×256 pixels in size. Examples of input images used in adversarial experiments are presented in Fig. 1 and Fig. 2.

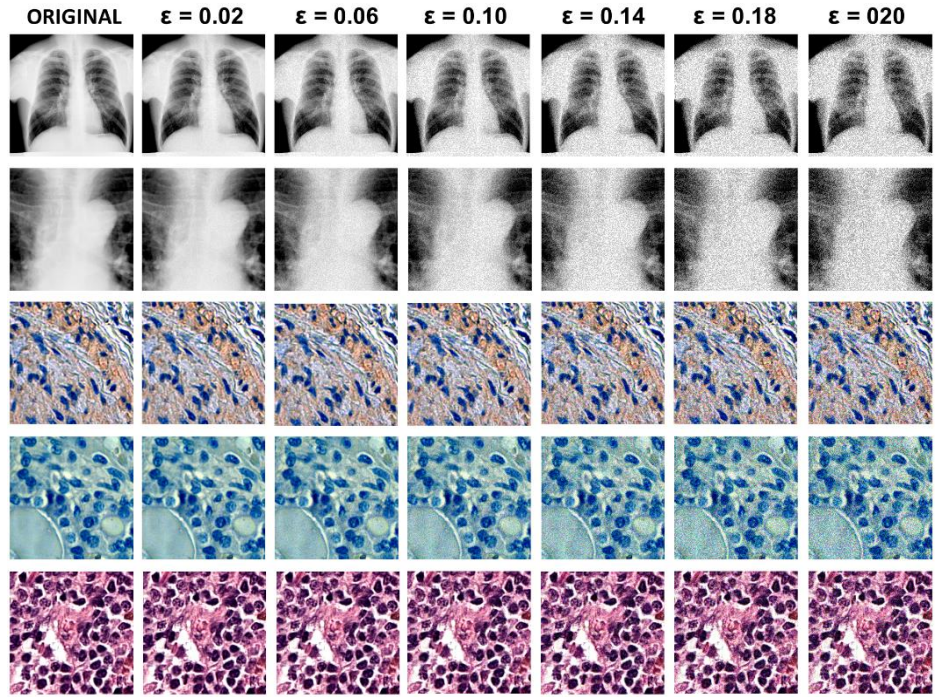

**Fig. 1.** Examples of original images and their adversarial versions with different amplitude ε.

## 3    Experimental Setup

The image data of each study group presented in Table 1 were split into the training and validation datasets in the proportion of 80/20 keeping same balance of class representatives. After that we trained several neural networks, one for each study group. The achieved classification scores are listed in last column of the Table 1. In order to reduce the variability of essential parameters, we used well known InceptionV3 architecture as a single model and AdamOptimizer training optimizer.

As soon as the training was completed, we started to examine adversarial attacks. Usually generation of adversarial images is performed by applying specific intensity modifications to original images for producing their adversarial versions. Depending on what information are needed to generate adversarial examples the algorithms are categorized into the white-box and black-box attacks. For white-box methods architecture of network and trained weights should be available. Furthermore, in some cases loss function and training optimizer are also required. For black-box attacks the necessary information include the type of images that classified by DNN. In this paper we study the white-box Projected Gradient Descent (PGD) attacks algorithm. This algorithm is of gradient-based type. That means that the source of perturbations is a gradient of some function taken from DNN that can be calculated rapidly with the help of back-propagation technique. The formula of PGD attack is given by:

$$x_{k+1} = x_k - lr * \text{Clip}_{x,\varepsilon}\left(\nabla_x\left(y_p(x_k)\right)\right), k \in [0, max\_iter - 1] \qquad (1)$$

In this equation the following three parameters are introduced: maximal amplitude of perturbation $\varepsilon$, number of executed iterations *max_iter* and the learning rate coefficient *lr*. In this work, we are focusing on examining the influence of $\varepsilon$ and *max_iter*t. The learning rate was not changed during the experiments and the same value of it was used for all classification tasks. Finally, we carry out the experiments in the following way:

(1) Define the set of values of $\varepsilon$ which is going to be examined.

(2) For every value of $\varepsilon$ and every image *x* in validation set of a classification task generate adversarial image $x_k$ by formula (1), where *k* iterates from 1 to *max_iter*.

(3) Save probabilities that were predicted by DNN for adversarial images. We don't need to store the images themselves because saving logits is enough to perform further analysis.

## 4    Results

The experimentation pipeline described above was executed for all image datasets listed in Table 1. Values of the amplitude of perturbation $\varepsilon$ were sampled evenly from the interval ranged from 0.02 to 0.20 with a step of 0.02 what is resulted in 10 repetitions of computational experiments. Examples of original images and their adversarial versions with different amplitude $\varepsilon$ are given in Fig. 1. Eight illustrative examples of successful attacks are shown in Fig. 2. In all these attacks the adversarial image examples were created using amplitude $\varepsilon$=0.02 and the number of iterations $N_I = 20$.

In order to assess the results of the attacks, we calculated the fraction of the images that have been misclassified. The "native" misclassifications made by DNN which are not related to attacks are not counted. We referred to this measure of success as the *Attack success rate*. At the first step we studied the dependency of the attack success rate on the perturbation amplitude $\varepsilon$. Motivation for looking onto this relation steams from the idea that permission to generate more perturbations with larger magnitude should lead to a higher attack rate. This is expected due to the nature of adversarial examples which is explained in [5]. Indeed, allowing the algorithm to search examples in a wider neighborhood gives it a greater chance to detect proper perturbations. The results summarized in form of corresponding plots of Fig. 3a suggest that the proposed idea is almost right: for 7 out of 8 datasets the expectation was confirmed.

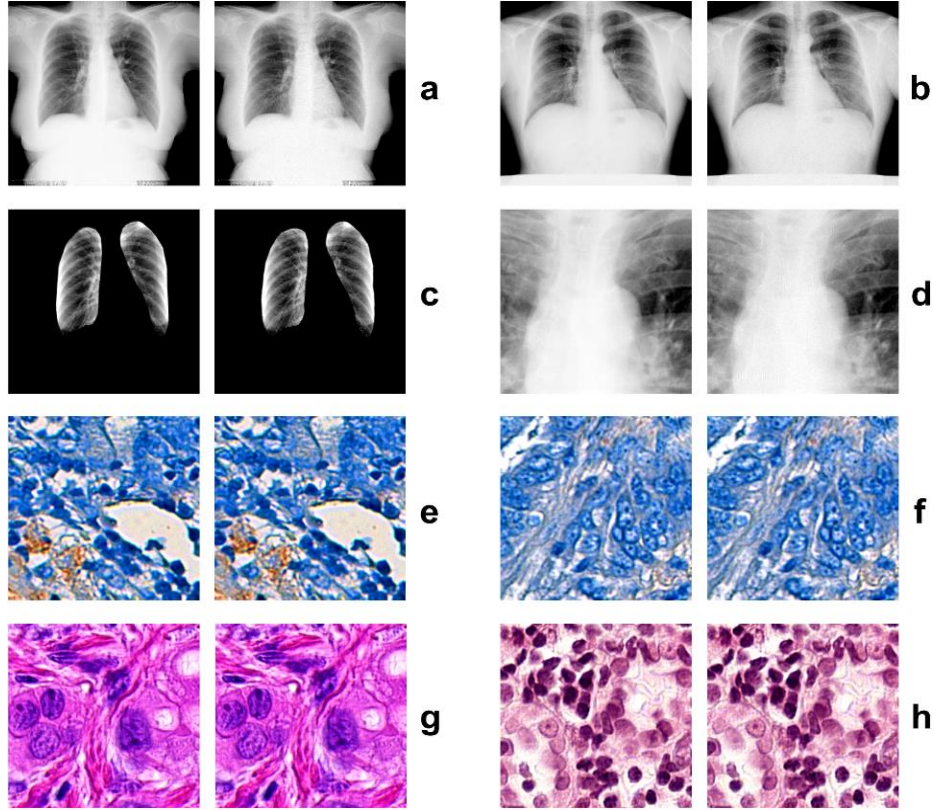

**Fig. 2.** Examples of successful attacks. Each image pair shows original image (left) and its adversarial version (on the right): (a) Aged person from the group 50-70 recognized as young one from the 20-35 group. (b) Young male subject from the group G1 aged 17-24 of classification task X-NR3 categorized to 42-80 group G3. (c) Tuberculosis patient misclassified to Norm. (d) Normal aorta misclassified as Pathology. (e) Normal Ovary tissue diagnosed as Cancerous. (f) Tissue of Thyroid Gland recognized as Ovary. (g) Breast cancer misclassified as Norm. (h) Fragment of image with normal tissue of breast was wrongly identified as being malignant.

It should be noted that the nature of adversarial examples does not prove the statement in a formal way. As our algorithm moves in direction of gradient with fixed *lr* there is no warranty that heavier perturbations would not "overstep" the desired class. This could be the reason that the plot corresponding to the X-AO dataset demonstrates sort of descending behavior.

For better understanding of significance of the number of iterations, we have considered dependency of attack success rate on it. In general, the step-by-step modification of input images should gradually increase the rate and asymptotically become a constant when the amount of perturbation is high enough and almost every pixel lies on defined boundary. In this study such an assumption was confirmed what is evident from Fig. 3b.

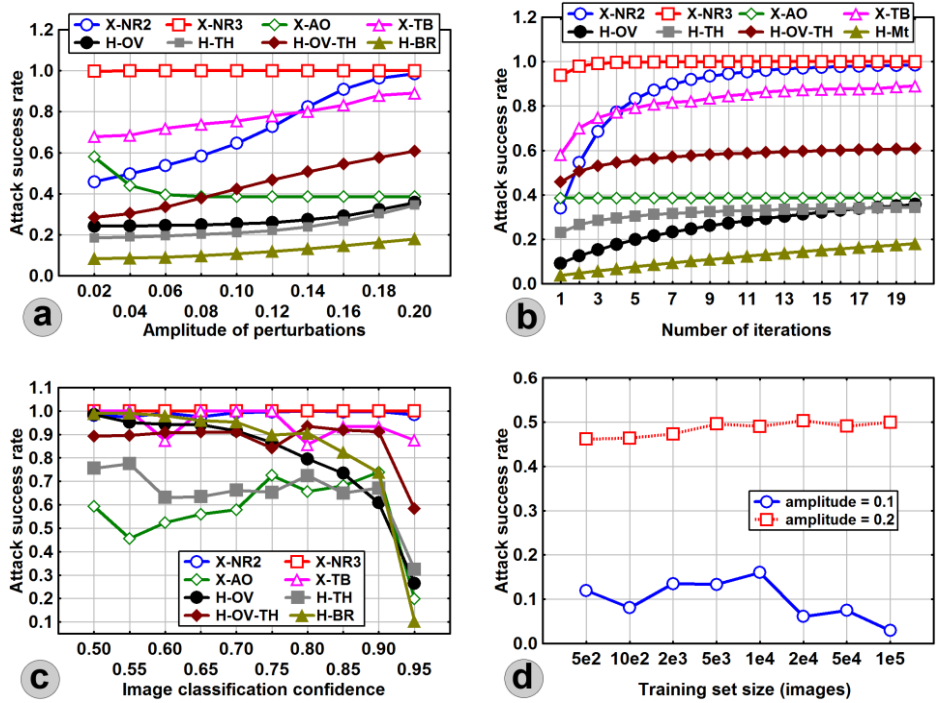

**Fig. 3.** Dependence of attack success rate on the amplitude of perturbations (a), the number of iterations (b), image classification confidence (c), and the size of image training set.

As it can be seen from the figure, in 6 out of 8 datasets the limit seems to be reached while for 2 remaining the slope of plots appears to be sufficiently high.

During the experimentation it has been noticed that there is certain relation between the probability, i.e., confidence of categorization of the original, unperturbed image to a specific class and the result obtained on the subsequent step of its attack. In order to discover whether this phenomenon is regular, we have performed additional experiment which consists of the following steps:

(1) Bining the probability range of [0.5, 1.0] into 10 equal intervals (bins).

(2) Categorization of each image into one of 10 bins using its maximal probability produced by neural network. For instance, in case the output probabilities of two-class classification task predicted as [0.81, 0.19] it should be categorized into bin [0.80, 0.85].

(3) Calculate attack success rate separately for each histogram bin.

Fig. 3c shows that the images failed to the bins which correspond to the highest probabilities are misclassified less often than the others. Furthermore, for image datasets H-OV and H-OV-TH this pattern is very prominent. Such a behavior could be caused by correlation between the image's probability and the size of class subspace. Namely, in case of high probability the classes' own neighborhoods should be larger.

Finally, we examined the links between the size of training image dataset and the attack success rate. To this end, we gradually increase the dataset size by way of sampling more and more images from H-BR, re-training the DNN, and attacking it again.

However, results shown in Fig. 3d gives us no reason to support such a conclusion on our image datasets and conditions of experimentation.

## 5      Conclusions

Results reported with this study allow drawing the following conclusions.

(1) An increase of the amplitude of perturbation in generating malicious adversarial images leads to a growth of the fraction of successful attacks for the majority of image types examined in this study. The only exception was demonstrated by X-ray images of the aorta with anterior rotation which represent sort of "geometrical" kind of pathological changes.

(2) Histology images tend to be less sensitive to the growth of amplitude of adversarial perturbations.

(3) Percentage of successful attacks is growing with an increase of the number of iterations of the algorithm of generating adversarial perturbations with an asymptotic stabilization.

(4) It was found that the success of adversarial attacks dropping dramatically when the original confidence of predicting image class exceeds 0.95.

(5) Dependence of percentage of successful attacks on the size of image training set was not confirmed.

## References

1. Litjens G., Kooi T., Bejnordi B., Setio A., Ciompi F., Ghafoorian M.: A survey on deep learning in medical image analysis. Medical Image Analysis 42, 60-88 (2017).
2. Ker J., Wang L., Rao J., Lim T.: Deep Learning Applications in Medical Image Analysis. IEEE Access 6, 9375-9389 (2018).
3. Szegedy C., Wojciech Z., Sutskever I., Bruna J., Dumitru E., Goodfellow I., Fergus R.: Intriguing properties of neural networks. International Conference on Learning Representations (ICLR) 2014, pp. 1-10. Springer, Banff (2014).
4. Goodfellow I., Shlens J., Szegedy C.: Explaining and harnessing adversarial examples. arXiv preprint arXiv:1412.6572v3 (2015).
5. Papernot N., McDaniel P., Jhay S., Fredriksonz M., Celik Z.B., Swamix A.: The Limitations of Deep Learning in Adversarial Settings. arXiv preprint arXiv:1511:07528v1 (2015).
6. Akhtar N., Mian A.S.: Threat of Adversarial Attacks on Deep Learning in Computer Vision. IEEE Access 6, 14410–14430 (2018).
7. Ozdag M.: Adversarial Attacks and Defenses Against Deep Neural Networks: A Survey. Procedia Computer Science 140, 152–161 (2018).
8. Madry A., Makelov A., Schmidt L., Tsipras D., Vladu A.: Towards Deep Learning Models Resistant to Adversarial Attacks. arXiv preprint arXiv:1706.06083v3 (2017).
9. Veta M., Heng Y.J., Stathonikos N. et. al.: Predicting breast tumor proliferation from whole-slide images. Medical Image Analysis 54, 111–121 (2019).